\documentclass{article}

 \PassOptionsToPackage{numbers}{natbib}

\usepackage[final]{neurips_2021}

\usepackage[utf8]{inputenc} 
\usepackage[T1]{fontenc}    
\usepackage{hyperref}       
\usepackage{url}            
\usepackage{booktabs}       
\usepackage{amsfonts}       
\usepackage{nicefrac}       
\usepackage{microtype}      
\usepackage{xcolor}         
\usepackage{graphicx}
\usepackage{wrapfig}
\usepackage{listings}
\usepackage{amsmath}
\usepackage{xcolor}
\usepackage{amsmath}

\usepackage{subcaption}

\usepackage{tikz}
\newcommand*\circled[1]{\tikz[baseline=(char.base)]{
            \node[shape=circle,draw,inner sep=2pt] (char) {#1};}}
            
\newcommand{\hush}[1]{}
            
\title{Practical Policy Optimization
\\ with Personalized Experimentation}

\author{%
   Mia Garrard, Hanson Wang, Ben Letham, Shaun Singh, Abbas Kazerouni, Sarah Tan, 
  \\ \textbf{Zehui Wang, Yin Huang, Yichun Hu, Chad Zhou, Norm Zhou, Eytan Bakshy}\\
  Meta, Menlo Park, CA\\ \\
  \small\texttt{mgarrard@meta.com}, 
  \small\texttt{hanson.wng@gmail.com},
  \small\texttt{bletham@meta.com}, \\
  \small\texttt{shaundsingh@gmail.com}, 
  \small\texttt{kazerouni@meta.com},
  \small\{\texttt{shftan, zehuiw.wang}\}\texttt{@gmail.com}, \\
  \small\texttt{maggiehuang@meta.com}, 
  \small\texttt{yh767@cornell.edu}, 
  \small\{\texttt{yuzhoubrother, nzhou, ebakshy}\}\texttt{@meta.com}\\
}

\begin{document}

\maketitle

\begin{abstract}
Many organizations measure treatment effects via an experimentation platform to evaluate the casual effect of product variations prior to full-scale deployment. However, standard experimentation platforms do not perform optimally for end user populations that exhibit {\em heterogeneous treatment effects} (HTEs). Here we present a personalized experimentation framework, {\em Personalized Experiments} (PEX), which optimizes treatment group assignment at the user level via HTE modeling and sequential decision policy optimization to optimize multiple short-term and long-term outcomes simultaneously. We describe an end-to-end workflow that has proven to be successful in practice and can be readily implemented using open-source software.

\end{abstract}
\section{Introduction}
Randomized experiments or ``A/B tests'' are a simple and routine method for evaluating potential improvements to internet services. Experimentation platforms abstract away the complexities of running such experiments from product owners resulting in higher accuracy and faster development time \cite{bakshy2014designing, xu2015infrastructure, colin2020netflix, kohavi2013online}. However, most A/B testing systems seek to estimate the {\em average treatment effect} (ATE) of a single product variation for all subjects, with the goal of selecting a single best variation. This methodology neglects heterogeneity in treatment response, which may be substantial in many settings.

Identifying heterogeneous treatment effects (HTEs) and identifying optimal policies given such heterogeneity has been of wide interest across medicine, political science, and technology \cite{wager2018estimation, kunzel2019metalearners, nie2021quasi, garcin2013personalized}. For many internet services, personalization can benefit recommender systems, user experience, resource utilization, ad targeting, and notifications \cite{garcin2013personalized, garcin2013pen, tougucc2020hybrid, argwal2014Ads}. However, training and deploying machine learning models and optimizing policies based on such models can be complex in practice, often requiring large teams of experts to fine tune its functioning.

We present a framework, {\em Personalized Experiments} (PEX), which provides a simple and robust method for real-world optimization of HTE-based policies. PEX employs individualized treatment effect (ITE) models combined with a simple decision policy to trade off multiple outcomes by adaptively assigning subjects to individualized treatments instead of assigning all individuals to the single treatment which does best on average.
To construct our decision policy, we create a linear utility function \cite{louviere2000,letham2019jmlr} that assigns a score to each treatment based on a set of tune-able parameters, and selects a treatment that maximizes expected utility. This allows for generalized multi-treatment, multi-outcome optimization of $n$ treatments and $m$ outcomes.  These policies are then optimized in a black-box fashion using multi-objective Bayesian optimization, which allows experimenters to target long-term outcomes that may or may not be well-modeled by the ITEs \cite{bakshy2018ae, daulton2021nehvi, letham2019jmlr}.

\section{A Three Phase Approach}
\begin{figure}[h!]
\centering
\includegraphics[width=13.5cm]{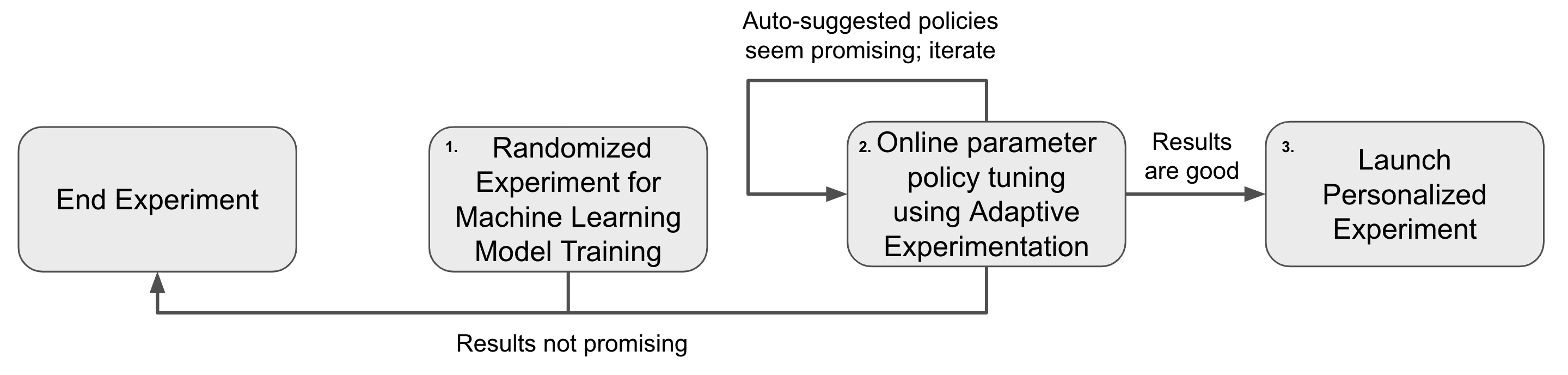}
\caption{Overview of the sequential decision points for experimenters in a PEX.}
\label{fig:3phase}
\end{figure}

To maximize product impact, PEX consists of three sequential phases:
\circled{1} ML model training with off-policy evaluation (OPE) of decision policies, 
\circled{2} online policy tuning,
and \circled{3} the launched personalized experiment, as opposed to the single experimentation step in traditional treatment effect testing. Below we describe each phase in detail. Due to noisy and non-stationary data, we cannot directly optimize for long-term outcomes, therefore we use a sequential approach to estimate proximal outcomes with HTE models and optimize via our decision policy. At each phase, experimenters decide how to proceed as shown in Figure \ref{fig:3phase}.  

\subsection{ML Model Training}
Experimenters setup a PEX via a {\em user interface} (UI) that is almost identical to the setup process for a traditional experimentation platform. The adopter identifies a set of $n$ treatments they wish to test (\textit{i.e.}, product variations), where the first treatment is typically a \textit{control treatment}. The policy will personalize the assignment of each user to one of these $n$ treatments. They also identify a collection of $m$ outcomes of interest, and flag the set of outcomes that should be maximized (or minimized). Adopters  select relevant features for HTE modeling via the UI; our platform automatically retrieves these features from a feature database via a client API, a common component of ML infrastructure at large internet companies \cite{Hazelwood2018FBData}. Subsequently, we train a collection of HTE models, generate candidate decision policies, and use off-policy evaluation to select policies for online evaluation.

\subsubsection{HTE Model Training}

The central component of the initial training phase is data collection to train the HTE models. During training, users are assigned to randomized treatments with equal probability. In addition to logging of treatment assignment and exposure, additional covariates ($x$) to be used in the HTE model are recorded at the time of first exposure. Combined with the outcomes for each user, this constitutes the training data for a collection of HTE models for each treatment and outcome combination. With $n$ treatment groups and $m$ outcomes, this yields $m (n - 1)$ distinct models, which we will refer to as $\hat{\tau}_{i, j}(x)$ with $2 \leq i \leq n$ for each treatment and $1 \leq j \leq m$ for each outcome. 

$\hat{\tau}_{i, j}(x)$ represents the model-based estimate of the conditional average treatment effect (CATE) given covariates $x$, for treatment $i$ compared to the control treatment, as measured on outcome $j$. To simplify notation, we include $i = 1$ as the control treatment and define $\hat{\tau}_{1, j} = 0$. 

The PEX framework feeds the training data to an internal AutoML platform to regularly train instances of the selected model type and abstract away the complexities of model training and upkeep. Most commonly, a HTE meta-learner \cite{kunzel2019metalearners} is trained to generate predictions for each outcome treatment pair. We have found that for most use cases, meta-learners based on gradient boosted regression trees with standard implementations such as \texttt{xgboost} \cite{xgboost2016} work well in practice.

\subsubsection{The Decision Policy}
\label{sec:decision-policy}

The decision policy is the component of PEX that converts HTE model predictions into treatment assignments. With a single outcome, as is common in the HTE literature, the most straightforward policy is to select the treatment with the maximum estimated treatment effect (i.e. $\text{argmax}_{1 \leq i \leq n} \hat{\tau}_{i, 1}(x)$). We discuss the generalization of such policies to multiple outcomes.

We assume that the desired trade-off between outcomes can be expressed using a \textit{linear utility function} \cite{louviere2000}. In particular, we assign a \textit{weight} $w_j \in \mathbb{R}$ to each outcome, which yields a linear utility function $u_i(x)$ for each treatment as follows:

\begin{align*}
  u_i(x) = \sum_{j=1}^{m}{w_j \cdot \hat{\tau}_{i, j}(x)}
\end{align*}

Since $\hat{\tau}$ are treatment effect estimates, the utility function is relative to the control treatment, where $u_1(x) = 0$ (as we have defined $\hat{\tau}_{1, j} = 0$ above.) Given this utility formulation, a basic policy that maximizes expected individual utility is to take the treatment corresponding to $\text{argmax}_{1 \leq i \leq n} u_i(x)$.

Empirically, we also find that introducing a per-treatment \textit{bias} term $b_i \in \mathbb{R}$ enables flexibility in shifting the policy towards or away from specific treatments and improves the overall Pareto frontier of achievable global tradeoffs (Figure \ref{fig:OPE}).

\begin{equation}
  u_i(x) = b_i + \sum_{j=1}^{m}{w_j \cdot \hat{\tau}_{i, j}(x)}
\end{equation}

An alternative interpretation of the $b_i$ term is to view it as a na\"ive regularizer of the treatment effect estimates $\hat{\tau}_{i, j}$ towards the (sample) average treatment effect of treatment $i$ for outcome $j$, notated as $\text{ATE}_{i, j}$. We can re-parameterize the bias by introducing a set of regularization parameters $\alpha_j \in [0, 1]$, which linearly interpolates between the ATE and estimate for each outcome $1 \leq j \leq m$: 

\begin{equation}
  u_i(x) = \sum_{j=1}^{m}{w_j [\alpha_j \cdot \text{ATE}_{i,j} + (1 - \alpha_j) \hat{\tau}_{i, j}(x)}]. \\
\end{equation}

If we assume that $\hat{\tau}_{i, j}(x)$ is a well-calibrated estimator (in the sense that $\mathbb{E}[\hat{\tau}_{i, j}(x)] = \text{ATE}_{i, j}$), the $\alpha_j$ adjustment can be viewed as a shrinkage estimator which reduces variance by a factor of $\alpha^2$. This can be helpful when combining outcomes with heterogeneous levels of measurement noise or error.

It can be seen that Equation (2) can be rearranged to the weights and biases formulation in Equation (1) with the substitution of $w'_j = w_j (1 - \alpha_j)$, $b'_i = \sum_{j=1}^{m}{w_j \alpha_j \cdot \text{ATE}_{i,j}}$. The more general formulation with $w_i, b_i \in \mathbb{R}$ is less restrictive and simplifies online policy deployment and tuning (Section \ref{sec:online-tuning}); appendix \ref{sec:bias-regularization} explores the exact restrictions in more detail.

We denote the policy under the vector of weights and biases ($w_j$, $b_i$) for a given set of treatments $\pi_{(w,b)}(x) := \text{argmax}_{1 \leq i \leq n} u_i(x)$; individuals are assigned to the treatment corresponding to $\pi_{(w,b)}(x)$. With the generalization of this linear decision policy to $n$ treatments and $m$ outcomes, the decision policy will have $m$ weights and $n$ biases. Without loss of generality, we can fix $w_1 = \{1, -1\}$ (1 for maximization, -1 for minimization) and $b_1 = 0$, scaling the other terms accordingly, producing a policy with $n + m - 2$ parameters. 

\hush{
Using an example where a product team is trying to optimize sending a notification in order to maximize login success, we  the generic linear decision policy for this case is as follows:

    \begin{align*}
            v_{test\_score}  = {w}_{login\_success}*(P(login\_success|A=1)-P(login\_success|A=0))\\ 
           + {w}_{notif\_cost}*(P(notif\_cost|A=1)-P(notif\_cost|A=0)) + {b}_{1},
    \end{align*}

where  ${w}_{login\_success}$ and ${w}_{notif\_cost}$ are the weights on the login success and notification cost outcomes respectively, $P({x}|{ y})$ is the HTE model prediction for outcome $x$ under treatment $y$, and ${b}_{1}$ is the constant bias term where a positive value shifts the users towards the treatment by increasing its score relative to other treatments.
}

\subsubsection{Off-policy evaluation}
The effect of deploying a particular personalized policy can be estimated via off-policy evaluation (OPE)~\cite{dudik2011drpe,bottou2013counterfactual} via subsampling or doubly-robust estimation, under the assumption that individuals maintain a fixed treatment over the period of interest. We note that in general a given policy may produce different treatment assignments if the covariates used by the ITE models change over time. A fixed treatment can be enforced via caching the treatment assignments upon initial exposure, in which case IPSW or standard contextual bandit policy evaluators such as  ~\citep{dudik2011drpe} could be reliably employed.  We find that in practice, even if such caching mechanisms are not in play, estimates from OPE that assume a fixed treatment regime tend to be biased but are ``directionally correct'' when compared with the online behavior which might be better characterized by a more complex MDP.


We use OPE to estimate the effects of the $n+m-2$-dimensional policy space implied by the utility model, and use multi-objective Bayesian optimization to perform offline optimization of these policies.  In particular, we utilize the qNParEGO algorithm \cite{daulton2020ehvi}, which maximizes noisy expected improvement \cite{letham2019constrained} under random scalarizations of the multiple outcomes. This algorithm scales well with the number of objectives and does not require decision makers to specify any additional information about their objectives.

\begin{figure}[h!]
\centering
\includegraphics[width=13.5cm]{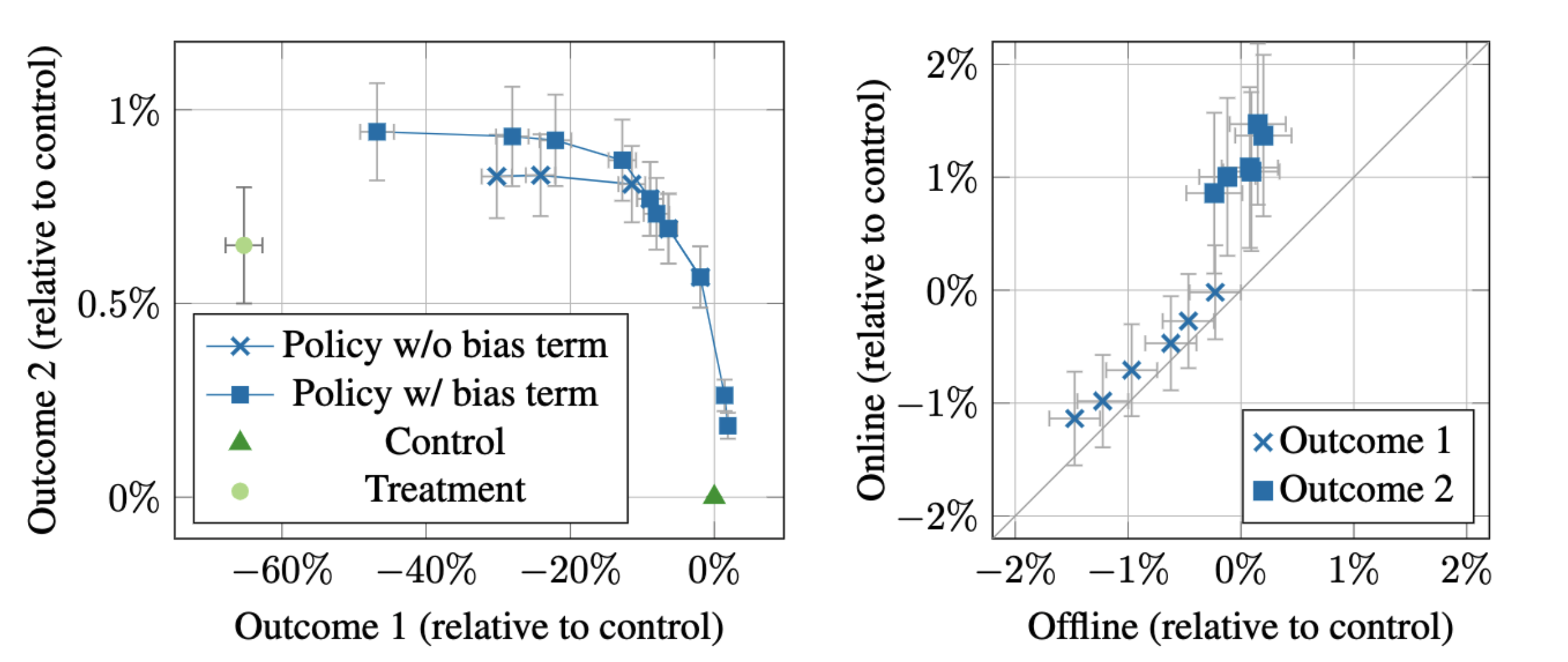}
\caption{\textbf{Left}: Example of the offline Pareto front for an experiment with $n=m=2$ outcomes and treatments (``treatment" and ``control"). Each square dot represents a generated decision policy and the associated OPE estimate for the outcome trade-off; error bars show 95\% confidence interval estimates. Both outcomes are competing maximization objectives (top right is better). \textbf{Right}: The association of the offline to online measurements for two metrics in an experiment.}
\label{fig:OPE}
\end{figure}

Figure \ref{fig:OPE} (Left) gives an example of the Pareto fronts achieved via Bayesian optimization using linear policies with and without bias terms.  One can see that personalized policies Pareto dominate a standard control or treatment assignment. Interestingly, one can see that that policies with bias terms Pareto dominate those without bias terms, and span a wider range of tradeoffs between what can be achieved strictly under the control or treatment conditions.

One limitation of this OPE approach is that only outcomes that were part of the HTE training data can be evaluated, which tend to be short-term, proximal outcomes. Ultimately we wish to optimize the policy for long-term outcomes. We use the offline Pareto front in combination with online experiments in order to directly optimize the policy for multiple long-term outcomes, as described in the next section.

\subsection{Online Policy Tuning} 
\label{sec:online-tuning}

Ultimately, our goal is to identify an optimal policy that maximizes some set of objectives that may or may not be accurately predicted by OPE. To accomplish this task, we leverage the offline OPE estimates to identify a small set of online experiments to conduct, after which point an experimenter may choose the policy that performs best, or perform a sequence of online or online/offline tuning experiments via Bayesian optimization~\cite{letham2019jmlr}. In particular, we generate an initial batch of online candidates from a subset of the offline Pareto optimal points, as in Figure \ref{fig:OPE} (Left). Given the ability to run $k$ online tests, we solve a subset selection problem to find the subset of $k$ offline Pareto points that maximize the hypervolume of the subsetted Pareto front. Beginning from this set of initial test policies, we proceed with an adaptive experiment to optimize for online outcomes in the decision policy parameter space. The offline optimization is used to find suitable parameter ranges for the online adaptive experiment.

During the online policy optimization  phase, subjects are assigned to decision policy candidates at random. We utilize Ax~\cite{bakshy2018ae} to deploy these experiments and optimize long-term metrics of interest via multi-objective Bayesian optimization.  We have found that in practice, OPE suffers from a difficult to predict bias which makes online testing of the candidate decision policies important for accurate experimentation. Figure \ref{fig:OPE} (Right) shows an example of such bias by associating the offline estimates to online measurements of two metrics in an experiment.

\subsection{Launching a Personalized Experiment}
Once the preferable decision policy from online Bayesian optimization is identified by experimenters, they can choose to run additional replication experiments or simply launch the personalized policy as the status quo.

To maintain model quality for the life of a launched PEX, we retain a small holdout (2\%-5\%) of end-users to continue receiving randomized treatments. This provides a data source for continuous model training, thus preventing staleness, and an effective backtest for experimenters to monitor the performance of a PEX policy versus randomized treatments. \hush{We also provide a configurable {\em stickiness} of model decisions, where experimenters select how frequently a user's treatment assignment should be assessed. This prevents poor end user experience by removing the risk of rapidly changing product features that could be frustrating.}

\section{Results and Discussion}

Here we presented a personalized experimentation platform, PEX, which allows for improving products by optimally personalizing treatment assignments at a user level. Our methodology supports multi-treatment, multi-objective optimization via Pareto optimization of a parameterized decision policy. With the construction of this decision policy, which includes both weight and bias parameters, we show an increase in Pareto hypervolume for our policies. 

To date, this system has enabled practically significant improvements at Meta, with respect to outcomes such as on-platform engagement, resource utilization, and harmful user behavior. Anecdotally, product engineers particularly appreciate PEX for its analysis tools, which provide insight into relevant user segments and potential trade-offs. Many PEX users have limited background in machine learning and/or causal inference, which speaks to the ease of use and robustness of our method. 

There are many areas of improvement that have become clear in our deployment. First, experimenters have a strong wish to understand policies generated by PEX. Methods from interpretable machine learning, such as policy distillation \cite{ czarnecki2019distilling, biggs2021model} or interpretable rule lists \cite{angelino2017learning} is a promising area of research that can help with demystifying the use of machine learning in treatment assignment. Second, the training procedure as described does not utilize data from the online model tuning to re-train the model.  Leveraging such observations could improve the quality of training data, and thus improve the overall policies with time.
Finally, we have found that short-term proxy outcomes tend to be much easier to predict compared with long-term outcomes that are of greater interest. Automating the selection of effective proxies and constructing effective surrogates continues to be an area of further research.

\bibliographystyle{plain}
\bibliography{refs}

\appendix

\section{Appendix}


\subsection{Bias term constraints resulting from the regularization formulation}
\label{sec:bias-regularization}

Continuing from Section \ref{sec:decision-policy}, recall the rearranged weight/bias formulation in terms of $w_j$ and $\alpha_j$:

\begin{subequations}
\begin{equation} \label{eq:w}
w'_j = w_j (1 - \alpha_j)
\end{equation}
\begin{equation} \label{eq:b}
b'_i = \sum_{j=1}^{m}{w_j \alpha_j \cdot \text{ATE}_{i,j}}
\end{equation}
\end{subequations}

Our goal is to characterize the set of all $w'_j$, $b'_i$ which have a solution $w_j \in \mathbb{R}$, $\alpha_j \in [0, 1]$ which satisfy the equations above.

Rearranging (\ref{eq:w}) to $w_j = w'_j / (1 - \alpha_j)$ and substituting into (\ref{eq:b}) yields:

\begin{equation*}
b'_i = \sum_{j=1}^{m}{\text{ATE}_{i,j} \cdot w'_j \frac{\alpha_j}{1 - \alpha_j}}
\end{equation*}

This can be expressed in matrix form as:

\begin{equation*}
\begin{pmatrix}
b'_1 \\
\vdots \\
b'_n
\end{pmatrix} = \begin{bmatrix}
    \text{ATE}_{1, 1} & \cdots & \text{ATE}_{1, m} \\
    \vdots & \ddots & \vdots \\
    \text{ATE}_{n, 1} & \cdots & \text{ATE}_{n, m}
\end{bmatrix} \begin{pmatrix}
w'_1 \cdot \frac{\alpha_1}{1 - \alpha_1} \\
\vdots \\
w'_m \cdot \frac{\alpha_m}{1 - \alpha_m}
\end{pmatrix}
\end{equation*}

Now, there is no restriction on the choice of $w'_j$ alone, so let us fix $w'_j \in \mathbb{R}$. Then, since $\frac{\alpha_j}{1 - \alpha_j} \in \mathbb{R}_{\geq 0}$, this restricts $b'$ to any \textit{conical combination} of the ATE columns adjusted by the sign of $w'_j$ (i.e. the \textit{conical hull}):

\begin{equation*}
\begin{pmatrix}
b'_1 \\
\vdots \\
b'_n
\end{pmatrix} \in \bigl\{ \sum_{i=1}^{m}{x_i \cdot \text{sgn}(w'_i) \cdot
\begin{pmatrix}
\text{ATE}_{1, i}\\
\vdots\\
\text{ATE}_{n, i}
\end{pmatrix}
},\, x_i \in \mathbb{R}_{\geq 0} \bigr\}
\end{equation*}

Note that as we have defined $\text{ATE}_{1, i} = 0$, this does already imply that $b'_1 = 0$ (though as mentioned previously, we can fix $b'_1 = 0$ without loss of generality). And, as the ATE matrix contains sample values computed from empirical data, we may assume that its rank is $\text{min}(n-1, m)$; when $m \geq n$, it is possible for the conical hull to span the entire remaining $\mathbb{R}^{n-1}$ space.

In practice, this tends to mean that most reasonable choices of weight and bias combinations do have an equivalent representation using the regularization interpretation. Consider the most common case of $n = 2$; depending on the alignments of signs between and $\text{ATE}_{2, j}$ and $w'_j$, $b'_2$ can be any value in either $\mathbb{R}_{\geq 0}$, $\mathbb{R}_{\leq 0}$, or even $\mathbb{R}$.

\end{document}